\documentclass[11pt]{article}

\usepackage[T1]{fontenc}
\usepackage[english]{babel}
\usepackage{amsmath,amssymb,amsthm}
\usepackage{amscd}
\usepackage{mathrsfs} 

\usepackage[shortlabels]{enumitem}
\usepackage{mathrsfs,euscript}
\usepackage{bm}

\RequirePackage[colorlinks,citecolor=blue,urlcolor=blue]{hyperref}

\usepackage{thmtools, thm-restate}
\usepackage[usenames,dvipsnames]{pstricks}
\usepackage{graphicx}
\usepackage{natbib}
\usepackage{newfloat}
\usepackage{booktabs}
\usepackage{algorithm}
\usepackage{algorithmic}

\usepackage[font=small]{caption}
\usepackage[font=scriptsize]{subcaption}

\usepackage[nameinlink,capitalize]{cleveref}

\usepackage{etoolbox}

\usepackage{microtype}
\usepackage{subcaption}
\usepackage{booktabs} 

\graphicspath{{./images/}}
\DeclareGraphicsExtensions{.eps,.pdf,.jpg,.png,}

\newcommand{\E}{\mathbb{E}}

\usepackage{footnote}
\makesavenoteenv{tabular}
\makesavenoteenv{table}

\makeatletter
\newcommand\footnoteref[1]{\protected@xdef\@thefnmark{\ref{#1}}\@footnotemark}
\makeatother

\providecommand{\scal}[2]{\left\langle{#1},{#2}\right\rangle}

\providecommand{\nor}[1]{\left\|{#1}\right\|}

\newcommand{\X}{{\mathcal X}}

\newcommand{\hh}{{\mathcal H}}

\newcommand{\R}{{\mathbb{R}}}

\newcommand{\eqals}[1]{\begin{align*}#1\end{align*}}

\title{Random Expert Distillation: Imitation Learning via Expert Policy Support Estimation}

\author{\small Ruohan Wang$^{1}$\quad Carlo Ciliberto$^1$  \quad Pierluigi Amadori$^{1}$ \quad Yiannis Demiris$^{1}$ \\ {\footnotesize\tt \{r.wang16,c.ciliberto,p.amadori,y.demiris\}@imperial.ac.uk} }
\begin{document}

\maketitle

\begin{abstract}
We consider the problem of imitation learning from a finite set of expert trajectories, without access to reinforcement signals.
The classical approach of extracting the expert's reward function via inverse reinforcement learning, followed by reinforcement learning is indirect and may be computationally expensive.
Recent generative adversarial methods based on matching the policy distribution between the expert and the agent could be unstable during training.
We propose a new framework for imitation learning by estimating the support of the expert policy to compute a fixed reward function, which allows us to re-frame imitation learning within the standard reinforcement learning setting.
We demonstrate the efficacy of our reward function on both discrete and continuous domains, achieving comparable or better performance than the state of the art under different reinforcement learning algorithms.
\end{abstract}

\section{Introduction}
\label{sec:intro}
\footnotetext[1]{Department of Electrical and
Electronic Engineering, Imperial College London, SW7 2BT, United Kingdom}

We consider a specific setting of imitation learning - the task of policy learning from expert demonstrations - in which the learner only has a finite number of expert trajectories without any further access to the expert. Two broad categories of approaches to this settings are behavioral cloning (BC) \cite{pomerleau1991efficient}, which directly learns a policy mapping from states to actions with supervised learning from expert trajectories; and inverse reinforcement learning (IRL) \cite{ng2000algorithms, abbeel2004apprenticeship}, which learns a policy via reinforcement learning, using a cost function extracted from expert trajectories.

Most notably, BC has been successfully applied to the task of autonomous driving \cite{bojarski2016end, bansal2018chauffeurnet}. Despite its simplicity, BC typically requires a large amount of training data to learn good policies, as it may suffer from compounding errors caused by covariate shift \cite{ross2010efficient, ross2011reduction}. BC is often used as a policy initialization step for further reinforcement learning \cite{nagabandi2018neural, rajeswaran2017learning}.

IRL estimates a cost function from expert trajectories and uses reinforcement learning to derive policies. As the cost function evaluates the quality of trajectories rather than that of individual actions, IRL avoids the problem of compounding errors. IRL is effective with a wide range of problems, from continuous control benchmarks in the Mujoco environment \cite{ho2016generative}, to robot footsteps planning \cite{ziebart2008maximum}.

Generative Adversarial Imitation Learning (GAIL) \cite{ho2016generative, baram2017end} connects IRL to the general framework of Generative Adversarial Networks (GANs) \cite{goodfellow2014generative}, and re-frames imitation learning as distribution matching between the expert policy and the learned policy via Jensen-Shannon Divergence. However, GAIL naturally inherits the challenges of GAN training, including possible training instability, and overfitting to training data \cite{arjovsky2017towards, brock2018large}. Similarly, generative moment matching imitation learning (GMMIL) considers distribution matching via maximum mean discrepancy. Both GAIL and GMMIL iteratively update the reward function and the learned policy during training.

In this paper, we propose imitation learning via expert policy support estimation, a new general framework for recovering a reward function from expert trajectories. We propose Random Expert Distillation (RED) by providing a connection between Random Network Distillation (RND) \cite{burda2018exploration} -- a method to design intrinsic rewards for RL exploration based on the "novelty" of states visited -- and support estimation ideas. Our method computes a fixed reward function from expert trajectories, and obviates the need for dynamic update of reward functions found in classical IRL, GAIL and GMMIL. Evaluating RED using different reinforcement learning algorithms on both discrete and continuous domains, we show that the proposed method achieves comparable or better performance than the state-of-arts methods on a variety of tasks, including a driving scenario (see Figure~\ref{fig:setup}). To the best of our knowledge, our method is the first to explore expert policy support estimation for imitation learning.

\section{Background}
We review the formulation of Markov Decision Process (MDP) in the context of which we consider imitation learning. We also review previous approaches to imitation learning and support estimation related to our proposed method.

~\newline\noindent{\bf Setting.} We consider an infinite-horizon discounted MDP, defined by the tuple $(S, A, P, r, p_0, \gamma)$, where $S$ is the set of states, $A$ the set of actions, $P :
S \times A \times S \rightarrow [0, 1]$ the transition probability, $r : S \times A \rightarrow \mathbb{R} $ the reward function, $p_0 : S \rightarrow [0, 1]$ the distribution over initial states, and $\gamma \in (0, 1)$ the discount factor. Let $\pi$ be a stochastic policy $\pi : S \times A \rightarrow [0, 1]$ with expected discounted reward $\E_{\pi}(r(s, a)) \triangleq \E(\sum_{t=0}^{\infty} \gamma^tr(s_t, a_t))$ where $s_0 \sim p_0$, $a_t \sim \pi(\cdot|s_t)$, and $s_{t+1} \sim P(\cdot |s_t, a_t)$ for $t \geq 0$. We denote $\pi_E$ the expert policy.

\subsection{Imitation Learning}
We briefly review the main methods for imitation learning:

~\newline\noindent{\em Behavioral Cloning (BC)} learns a control policy $\pi : S \rightarrow A$ directly from expert trajectories via supervised learning. Despite its simplicity, BC is prone to compounding errors: small mistakes in the policy cause the agent to deviate from the state distribution seen during training, making future mistakes more likely. Mistakes accumulate, leading to eventual catastrophic errors \cite{ross2011reduction}. While several strategies have been proposed to address this \cite{ross2010efficient, sun2017deeply}, they often require access to the expert policy during the entire training process, rather than a finite set of expert trajectories. BC is commonly used to initialize control policies for reinforcement learning \cite{rajeswaran2017learning, nagabandi2018neural}. 

~\newline\noindent{\em Inverse Reinforcement Learning (IRL)} models the expert trajectories with a Boltzmann distribution \cite{ziebart2008maximum}, where the likelihood of a trajectory is defined as:
\begin{equation}
    p_{\theta}(\tau) = \frac{1}{Z}\exp(-c_{\theta}(\tau))
\end{equation}
where $\tau=\{s_1, a_1, s_2, a_2 ...\}$ is a trajectory, $c_\theta(\tau)=\sum_t c_\theta(s_t, a_t)$ a learned cost function parametrized by $\theta$, and the partition function $Z\triangleq \int \exp(-c_\theta(\tau))d(\tau)$ the integral over all trajectories consistent with transition function of the MDP. The main computational challenge of IRL is the efficient estimation the partition function $Z$. IRL algorithms typically optimize the cost function by alternating between updating the cost function and learning an optimal policy with respect to the current cost function with reinforcement learning \cite{abbeel2004apprenticeship, ng2000algorithms}.

~\newline\noindent{\em Imitation Learning via Distribution Matching} frames imitation learning as distribution matching between the distribution of state-action pairs of the expert policy and that of the learned policy. In \cite{ho2016generative, finn2016connection}, the authors connect IRL to distribution matching via Generative Adversarial Networks (GANs) \cite{goodfellow2014generative}. Known as Generative Adversarial Imitation Learning (GAIL), the method imitates an expert policy by formulating the following minimax game:
\begin{equation}
    \min_{\pi} \max_{D \in (0, 1)}~ \E_{\pi} (\log D(s, a)) + \E_{\pi_E}(\log (1-D(s, a))) - \lambda H(\pi) 
\end{equation}
where the expectations $\E_{\pi}$ and $\E_{\pi_E}$ denote the joint distributions over state-action pairs of the learned policy and the expert policy, respectively, and  $H(\pi) \triangleq \E_{\pi}(-\log \pi(a|s))$ is the entropy of the learned policy. GAIL has been successfully applied to various control tasks in the Mujoco environment \cite{ho2016generative, baram2017end}. However, GAIL inherits the challenges of GANs, including possible training instability such as vanishing or exploding gradients, as well as overfitting to training data \cite{arjovsky2017towards, brock2018large}. While numerous theoretical and practical techniques have been proposed to improve GANs (e.g \cite{arjovsky2017wasserstein, salimans2016improved}), a large-scale study of GANs show that many GAN algorithms and architectures achieve similar performance with sufficient hyperparameter tuning and random restarts, and no algorithm or network architecture stands out as the clear winner on all tasks \cite{lucic2018gans}.

Similar to GAIL, \cite{kim2018imitation} proposed generative moment matching imitation learning (GMMIL) by minimizing the maximum mean discrepancy between the expert policy and the learned policy. Though GMMIL avoids the difficult minimax game, the cost of each reward function evaluation grows linearly with the amount of training data, which makes scaling the method to large dataset potentially difficult. In addition, we demonstrate in our experiments that GMMIL may fail to estimate the appropriate reward functions.

\subsection{Support Estimation with Kernel Methods}\label{sec:support-estimation-kernel}

As we will motivate in detail in Sec.~\ref{sec:algorithm}, we argue that estimating the support of the expert policy can lead to good reward functions for imitation learning. In this section we review one of the most well-established approaches to support estimation of a distribution from a finite number of i.i.d. samples, which relies on a kernelized version of principal component analysis \cite{scholkopf1998nonlinear}. The idea is to leverage the {\em Separating Property} \cite{de2014universally} of suitable reproducing kernel Hilbert spaces, which guarantees the covariance operator of the embedded data to precisely capture the geometrical properties of the support of the underlying distribution. This allows us to derive a test function that is zero exclusively on points belonging to the support of the distribution.

Formally, let $\X\subseteq\R^d$ be a set (in our setting $\X = S \times A$ is the joint state-action space) and let $k:\X\times\X\to\R$ be a positive definite kernel with associated reproducing kernel Hilbert space (RKHS) $\hh$ \cite{aronszajn1950theory} and feature map $\phi:\X\to\hh$, such that $k(x,x') = \scal{\phi(x)}{\phi(x')}_\hh$ for any $x,x'\in\X$. For any set $U\subseteq\X$, denote by $\Phi(U) = \{\phi(x)~|~x\in U\}$ and $\overline{\Phi(U)}$ the closure of its span in $\hh$. The separating property guarantees that for any closed subset $U$ of $\X$, $\Phi(U) = \Phi(\X) \cap \overline{\Phi(U)}$. In other words, the separating property ensures that 
\begin{equation}
\label{eq:sep_cond}
    x\in U ~~\Longleftrightarrow~~ \phi(x)\in\overline{\Phi(U)}.
\end{equation}
As shown in \cite{de2014universally}, several kernels allow the separating property to hold, including the popular Gaussian kernel $k(x,x') = \exp(-\nor{x-x'}/\sigma)$ for any $\sigma>0$.


The separating property suggests a natural strategy to test whether a point $x\in\X$ belongs to a closed set $U$. Let $P_U:\hh\to\hh$ be the orthogonal projection operator onto $\overline{\Phi(U)}$ (which is a linear operator since $\overline{\Phi(U)}$ is a linear space), we have
\eqals{
    x\in U ~~\Longleftrightarrow~~ \nor{(I - P_U)\phi(x)}_\hh = 0,
}
since Eq. (\ref{eq:sep_cond}) corresponds to $\phi(x) = P_U \phi(x)$, or equivalently $(I-P_U)\phi(x) = 0$. 

We can leverage the machinery introduced above to estimate the support ${\textrm{supp}}(\pi)\subseteq\X$ of a probability distribution $\pi$ on $\X$. We observe that the covariance operator $C_\pi = \E_\pi[\phi(x)\phi(x)^\top]$ encodes sufficient information to recover the orthogonal projector $P_{\textrm{supp}(\pi)}$ (denoted $P_\pi$ in the following for simplicity). More precisely, let $C_\pi^\dagger$ denote the pseudoinverse of $C_\pi$. A direct consequence of the separating property is that $P_\pi = C_\pi^\dagger C_\pi$ \cite{de2014universally}.

When $\pi$ is unknown and observed only through a finite number of $N$ i.i.d examples $\{x_i\}_{i=1}^N$, it is impossible to obtain $C_\pi$ (and thus $P_{\pi}$) exactly. A natural strategy is to consider the {\em empirical covariance} operator $\hat C = \frac{1}{N}\sum_{i=1}^N \phi(x_i)\phi(x_i)^\top$ as a proxy of the ideal one. Then, the projector is estimated as $\hat P = \hat C_m^\dagger \hat C_m$, where $C_m$ is the operator comprising the $m\leq n$ principal component directions of $C$, where $m$ is a hyperparameter to control the stability of the pseudoinverse when computing $\hat P$. We can then test whether a point $x\in \X$ belongs to ${\textrm{supp}}(\pi)$, by determining if
\begin{equation}\label{eq:supp_score}
    \nor{(I-\hat P)\phi(x)}_\hh^2 = \scal{\phi(x)}{(I-\hat P)\phi(x)}_\hh = k(x,x) - \scal{\phi(x)}{\hat P\phi(x)}_\hh,
\end{equation}
is greater than zero (in practice a threshold $\tau>0$ is introduced). Note that we have used the fact that $\hat P^\top\hat P = \hat P$, since $\hat P$ is an orthogonal projector. Although $\hat C$ and $\hat P$ are operators between possibly infinite dimensional spaces, we have for any $x\in\X$ \citep[see][]{scholkopf1998nonlinear}
\eqals{
    \scal{\phi(x)}{\hat P \phi(x)}_\hh = K_x^\top K_m^\dagger K_x,
}
where $K\in\R^{n\times n}$ is the empirical kernel matrix of the training examples, with entries $K_{ij} = k(x_i,x_j)$. Here, $K_m$ denotes the matrix obtained by performing PCA on $K$ and taking the first $m\leq n$ principal directions and $K_x\in\R^n$ is the vector with entries $(K_x)_i=k(x_i,x)$. Therefore $\langle{\phi(x)},{\hat P \phi(x)}\rangle_\hh$ can be computed efficiently in practice. 

The theoretical properties of the strategy described above have been thoroughly investigated in the previous literature \cite{de2014universally,rudi2017regularized}. In particular, it has been shown that the Hausdorff distance between $\textrm{supp}(\pi)$ and the set induced by $\hat P$, tends to zero when $n\to+\infty$, provided that the number of principal components $m$ grows with the number of training examples. Moreover, it has been shown that the support estimator enjoys fast learning rate under standard regularity assumptions. 

\section{Imitation Learning via Expert Policy Support Estimation}\label{sec:algorithm}

Formally, we are interested in extracting a reward function $\hat r(s, a)$ from a finite set of trajectories $D=\{\tau_i\}_{i=1}^N$ produced by an expert policy $\pi_E$ within a MDP environment. Here each $\tau_i$ is a trajectory of state-action pairs of the form $\tau_i=\{s_1, a_1, s_2, a_2, ..., s_T, a_T\}$. Assuming that the expert trajectories are consistent with some unknown reward function $r^*(s, a)$, our goal is for a policy learned by applying RL to $\hat r(s, a)$, to achieve good performance when evaluated against $r^*(s, a)$ (the standard evaluation strategy for imitation learning). In contrast with traditional imitation learning, we do not explicitly aim to match $\pi_E$ exactly.

We motivate our approach with a thought experiment. Given a discrete action space and a deterministic expert such that $a_s^*=\pi_E(s)$, the resulting  policy is a Dirac's delta at each state $s\in S$, for all $(s, a)$. Consider the reward function
\begin{equation}
    r_E(s, a)=\begin{cases}
               1 \text{ if } (s, a) \in \textrm{supp}(\pi_E)\\
               0 \text{ if } (s, a) \notin \textrm{supp}(\pi_E)\\
            \end{cases}
\end{equation}
which corresponds to the indicator function of $\textrm{supp}(\pi_E)$. It follows that a RL agent trained with this reward function would be able to imitate the expert exactly, since the discrete action space allows the random exploration of the agent to discover optimal actions at each state. In practice, the expert policy is unknown and only a finite number of trajectories sampled according to $\pi_E$ are available. In these context we can use support estimation techniques to construct a reward function $\hat r$. 

For continuous action domains, sparse reward such as $r_E$ is unlikely to work since it is improbable for the agent, via random exploration, to discover the optimal actions, while all other actions are considered equally bad. Instead, since support estimation produces a score with Eq. (\ref{eq:supp_score}) for testing each input, we may directly use that score to construct the reward function $\hat r$. The rationale is that actions similar to that of the expert in any given state would still receive high scores from support estimation, allowing the RL agent to discover those actions during exploration.

Based on the motivation above, we hypothesize that support estimation of the expert policy's state-action distribution provides a viable reward function for imitation learning. Intuitively, the reward function encourages the RL agent to behave similarly as the expert at each state and remain on the estimated support of the expert policy. Further, this allows us to compute a fixed reward function based on the expert trajectories and re-frame imitation learning in the standard context of reinforcement learning.

We note that for stochastic experts, the support of $\pi_E$ might coincide with the whole space $S\times A$. Support estimation would hence produce an uninformative, flat reward function $r_E$ when given an infinite amount of training data. However, we argue that an infinite amount of training data should allow BC to successfully imitate the expert, bypassing the intermediate step of extracting a reward function from the expert trajectories.

\subsection{Practical Support Estimation}\label{sec:support-estimation-rnd-intuition}

Following the strategy introduced in Sec.~\ref{sec:support-estimation-kernel}, we consider a novel approach to support estimation. Our method takes inspiration from kernel methods but applies to other models such as neural networks.

Let $\hh$ be a RKHS and $f\in\hh$ a function parametrized by $\theta\in\Theta$. Consider the regression problem that admits a minimizer on $\hh$
\eqals{
    \inf_{f\in\hh} \int (f_\theta(x) - f(x))^2 ~d\pi(x),
}
The minimal $\nor{\cdot}_\hh$ solution corresponds to $f_{\pi,\theta} = C_\pi^\dagger C_\pi f_\theta = P_\pi f_\theta$, the orthogonal projection of $f_\theta$ onto the range of $C_\pi$ \citep[see e.g][]{engl1996regularization}. For any $x\in\X$ we have
\eqals{
    f_\theta(x) - f_{\pi,\theta}(x)= \scal{f_{\theta}}{(I-P_\pi)\phi(x)}_\hh.
}
In particular, if $x\in\textrm{supp}(\pi)$, we have $(I-P_\pi)\phi(x) = 0$ and hence $f_\theta(x) -f_{\pi,\theta}(x) = 0$. The converse is unfortunately not necessarily true: for a point $x\not\in \textrm{supp}(\pi)$, $(I-P_\pi)\phi(x)$ may still be orthogonal to $f_\theta$ and thus $f_\theta(x) - f_{\pi,\theta}(x) = 0$. 

A strategy to circumvent this issue is to consider multiple $f_\theta$, and then average their squared errors $(f_\theta - f_{\pi,\theta})^2$. The rationale is that if $x\not\in S_\pi$, by spanning different directions in $\hh$, at least one of the $f_\theta$ will not be orthogonal to $(I-P_\pi)\phi(x)$, which leads to a non-zero value when compared against $f_{\pi,\theta}(x)$. In particular, if we sample $\theta$ according to a probability distribution over $\Theta$, we have
\eqals{
    \E_\theta (f_\theta - f_{\pi,\theta})^2 & = \E_\theta \scal{(I-P_\pi)\phi(x)}{f_\theta f_\theta^\top (I-P_\pi)\phi(x)}_\hh \\
    & = \scal{(I-P_\pi)\phi(x)}{\Sigma (I-P_\pi)\phi(x)}_\hh,
}
where $\Sigma = \E_\theta [f_\theta f_\theta^\top]$ is the covariance of the random functions $f_\theta$ generated from sampling the parameters $\theta$. Ideally, we would like to sample $\theta$ such that $\Sigma = I$, which allows us to recover the support estimator $\nor{(I-P_\pi)\phi(x)}_\hh$ introduced in Sec.~\ref{sec:support-estimation-kernel}. However, it is not always clear how to sample $f_\theta$ so that its covariance corresponds to the identity in practice. In this sense, this approach does not offer the same theoretical guarantees as the kernelized method reviewed in Sec.~\ref{sec:support-estimation-kernel}.

The discussion above provides us with a general strategy for support estimation. Consider a hypotheses space of functions $f_\theta$ parametrized by $\theta\in\Theta$ (e.g. neural networks). We can sample $\theta_1,\dots,\theta_K$ functions according to a distribution over $\Theta$. Given a training dataset $\{x_i\}_{i=1}^N$ sampled from $\pi$, we solve the $K$ regression problems 
\begin{equation}
\label{eq:prac_supp}
    \hat\theta_k = \min_{\theta\in\Theta} \frac{1}{N} \sum_{i=1}^N (f_\theta(x_i) - f_{\theta_k}(x_i))^2,    
\end{equation}
for every $k=1,\dots,K$. Then for any $x\in\X$, we can test whether it belongs to the support of $\pi$ by checking whether 
\begin{equation}
\label{eq:supp_test}
    \frac{1}{K}\sum_{k=1}^K (f_{\hat\theta_k}(x) - f_{\theta_k}(x))^2,   
\end{equation}
is larger than a prescribed threshold. As $K$ and $N$ grow larger, the estimate above will approximate increasingly well the desired quantity $\E_\theta (f_{\pi,\theta}(x)-f_\theta(x))^2$. 

\subsection{Reward Function with Random Network Distillation}
We may interpret the recent method of Random Network Distillation (RND) \cite{burda2018exploration} as approximate support estimation. Formally, RND sets up a randomly generated prediction problem involving two neural networks: a fixed and randomly initialized target network $f_{\theta}$ which sets the prediction problem, and a predictor network $f_{\hat{\theta}}$ trained on the state-action pairs of the expert trajectories. $f_{\theta}$ takes an observation to an embedding $f_{\theta} : S\times A \rightarrow \mathbb{R}^K$ and similarly so for $f_{\hat{\theta}}: S\times A \rightarrow \mathbb{R}^K$, which is analogous to the $K$ prediction problems defined in Eq. (\ref{eq:prac_supp}). The predictor network is trained to minimize the expected mean square error by gradient descent $L(s, a) = ||f_{\hat{\theta}}(s, a) - f_{\theta}(s, a)||_2^2$ with respect to its parameters $\hat{\theta}$. After training, the score of a state-action pair belonging to the support of $\pi_E$ is predicted by $L(s, a)$, analogous to Eq. (\ref{eq:supp_test}). One concern to RND is that a sufficiently powerful optimization algorithm may recover $f_{\theta}$ perfectly, causing $L(s, a)$ to be zero everywhere. Our experiments confirm the observations in \cite{burda2018exploration} that standard gradient-based methods do not behave in this undesirable way.

With the prediction error $L(s, a)$ as the estimated score of $(s, a)$ belonging to the support of $\pi_E$ , we propose a reward function that has worked well in practice,
\begin{equation}
\label{eq:supp_rew}
    r(s, a) = \exp(-\sigma_1 L(s, a))
\end{equation}
where $\sigma_1$ is a hyperparameter. As $L(s, a)$ is positive, $r(s, a) \in [0, 1]$. We choose $\sigma_1$ such that $r(s, a)$ from expert demonstrations are mostly close to 1.

\subsection{Terminal Reward Heuristic}
For certain tasks, there are highly undesirable terminal states (e.g. car crashes in autonomous driving). In such contexts, We further introduce a terminal reward heuristic to penalize the RL agent from ending an episode far from the estimated support of the expert policy. Specifically, let $\bar{r} = -\frac{1}{N}\sum_{i=1}^N r(s_i, a_i)$ be the average reward computed using expert trajectories, and $r(s_T, a_T)$ the reward of the final state of an episode, we define
\begin{equation}
\label{eq:term_rew}
    r_{term}=\left\{\begin{array}{cl}
               -\sigma_2 \bar{r} & \text{ if }  \sigma_3\bar{r} > r(s_T, a_T) \\
               0 & \text{ otherwise }\\
            \end{array}
            \right.
\end{equation}
where $\sigma_2, \sigma_3$ are hyperparameters. We apply the heuristic to an autonomous driving task in the experiment, and demonstrate that it corresponds nicely with dangerous driving situations and encourages RL agents to avoid them. We note that the heuristic is not needed for all other tasks considered in the experiments.

\begin{algorithm}[t]
   \caption{{\sc Random Expert Distillation}}
   \label{alg:red}
\begin{algorithmic}
   \STATE {\bfseries Input:} data $D=\{(s_i, a_i)\}_{i=1}^N$, $\Theta$ function models, initial policy $\pi_0$.
   
   \STATE~
   

   \STATE Randomly sample $\theta\in\Theta$
   \STATE $\hat\theta=${\sc Minimize}$(f_{\hat{\theta}},f_{\theta},D)$
   \STATE $r(\cdot) = \exp(-\sigma_1\|f_{\hat{\theta}}(\cdot)-f_{\theta}(\cdot)\|_2^2)$
   \STATE Compute $r_{\textrm{term}}$ from Eq. (\ref{eq:term_rew})
   \STATE $\pi=${\sc ReinforcementLearning}$(r,r_{\textrm{term}},\pi_0)$.
   
   \STATE~
   \STATE {\bfseries Return:} $\pi$.
\end{algorithmic}
\end{algorithm}

\subsection{Random Expert Distillation}
We present our algorithm Random Expert Distillation (RED) in Algorithm \ref{alg:red}. The algorithm computes the estimated support of the expert policy and generates a fixed reward function according to Eq. (\ref{eq:supp_rew}). The reward function is used in RL for imitation learning. As we report in the experiments, a random initial policy $\pi_0$ is sufficient for the majority of tasks considered in the experiments, while the remaining tasks requires initialization via BC. We will further discuss this limitation of our method in the results.

\subsection{Alternative to RND}
AutoEncoder (AE) networks \cite{vincent2008extracting} set up a prediction problem of $\min_\theta ||f_{\theta}(x)-x||_2^2$. AE behaves similarly to RND in the sense that it also yields low prediction errors for data similar to the training set. AE could be also seen as approximate support estimation in the context of Eq. (\ref{eq:supp_test}), by replacing randomly generated $f_{\theta_k}$ with identity functions on each dimension of the input. Specifically, AE sets up K prediction problems
\begin{equation*}
\min_{\theta} \frac{1}{N} \sum_{i=1}^N (f_\theta(x_i) - I_k(x_i))^2   
\end{equation*}
for $k=1,\dots,K$ where $I_k(x)=x_k$ and $K$ is the input size. However, as discussed in \cite{burda2018exploration}, AE with a bottleneck layer may suffer from input stochasticity or model misspecification, which are two sources of prediction errors undesirable for estimating the similarity of input to the training data. Instead, we have found empirically that overparametrized AEs with a $\ell_2$ regularization term prevent the trivial solution of an identity function, allow easier fitting of training data, and may be used in place of RND. We include AE in our experiments to demonstrate that support estimation in general appears to be a viable strategy for imitation learning.

\begin{figure*}[t]
    \centering
    \begin{subfigure}[t]{.24\textwidth}
        \includegraphics[width=\textwidth]{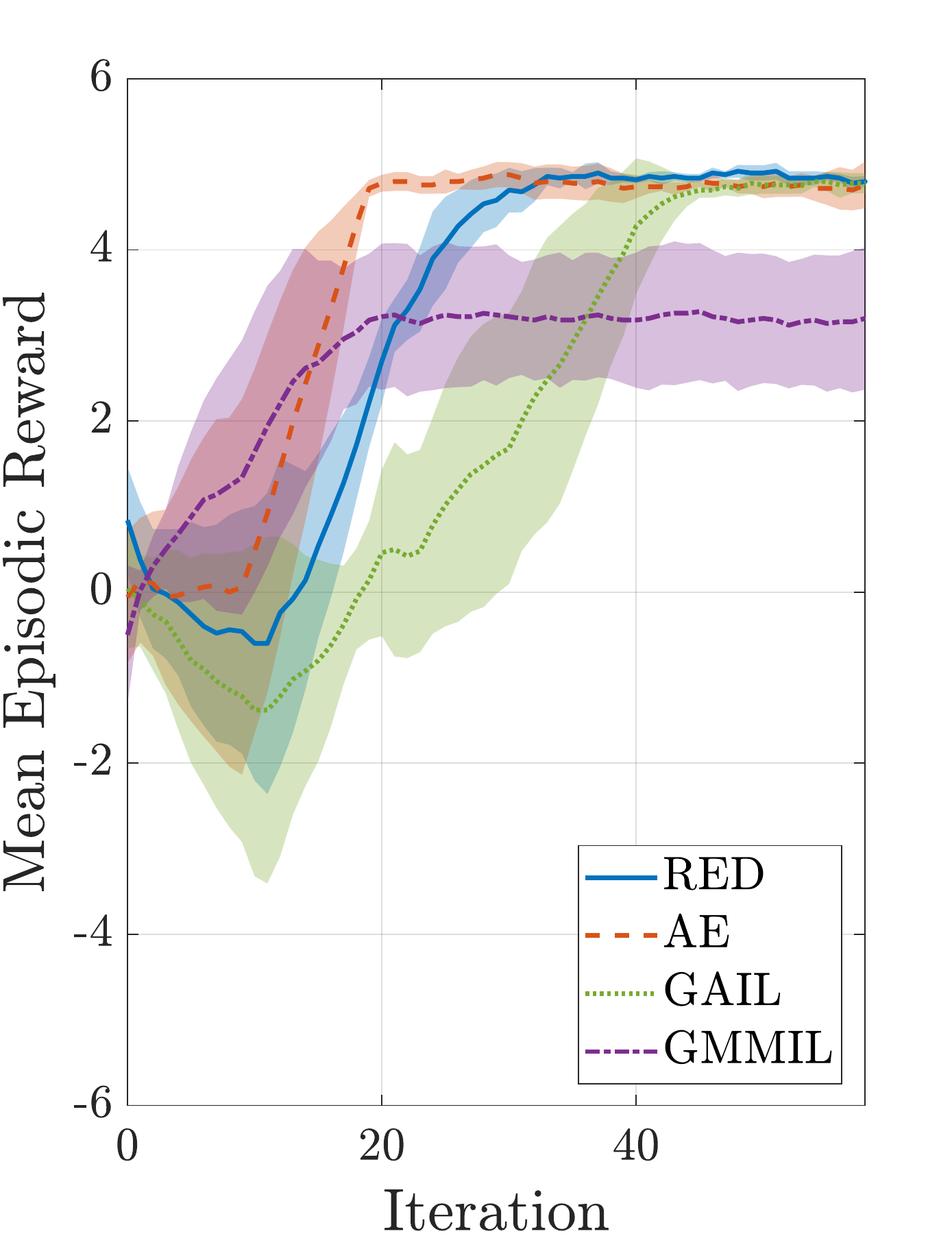}
        \caption{n = 5}
        \label{fig:toy_5}
    \end{subfigure}
    \begin{subfigure}[t]{.24\textwidth}
        \includegraphics[width=\textwidth]{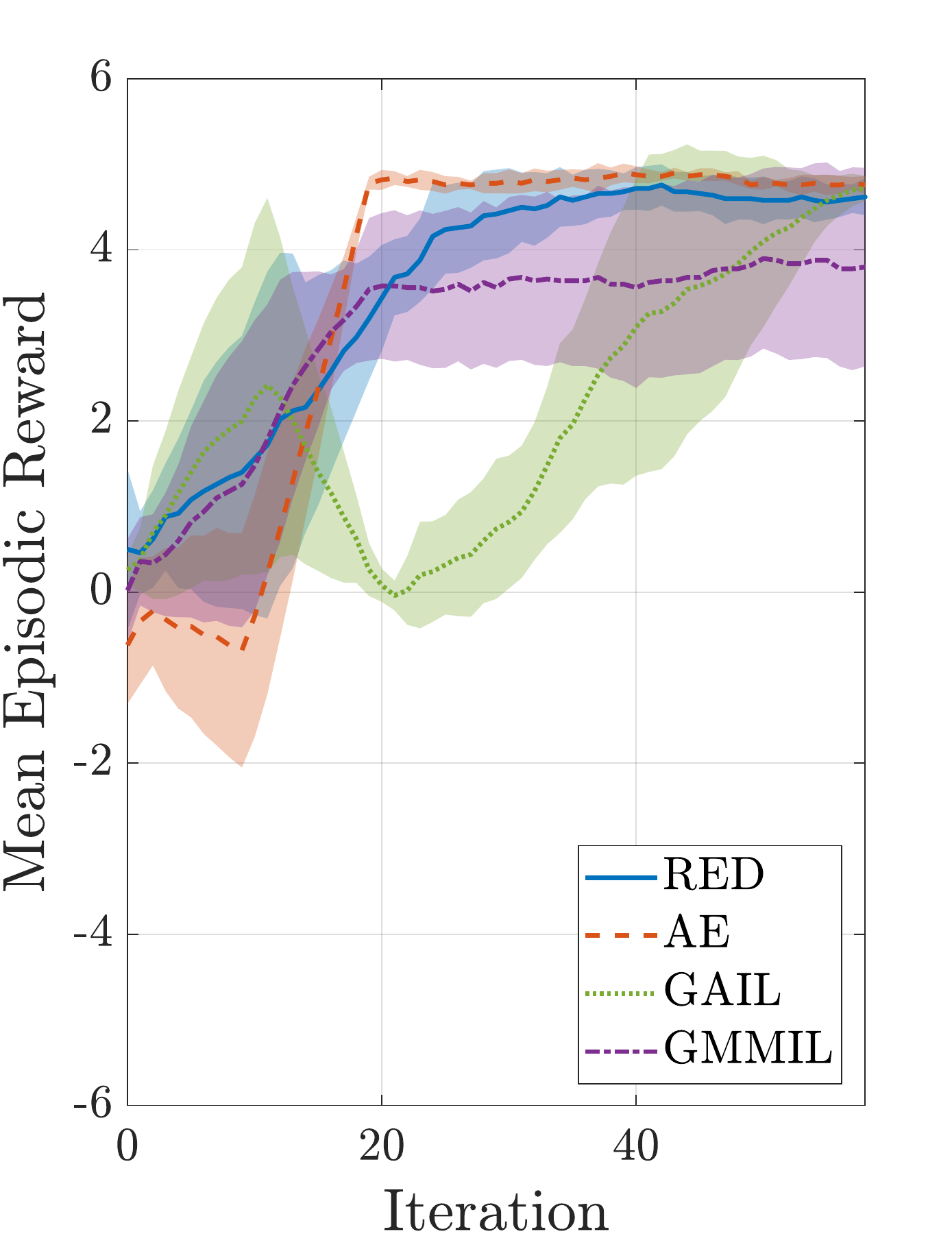}
        \caption{n = 10}
        \label{fig:toy_10}
    \end{subfigure}
    \begin{subfigure}[t]{.24\textwidth}
        \includegraphics[width=\textwidth]{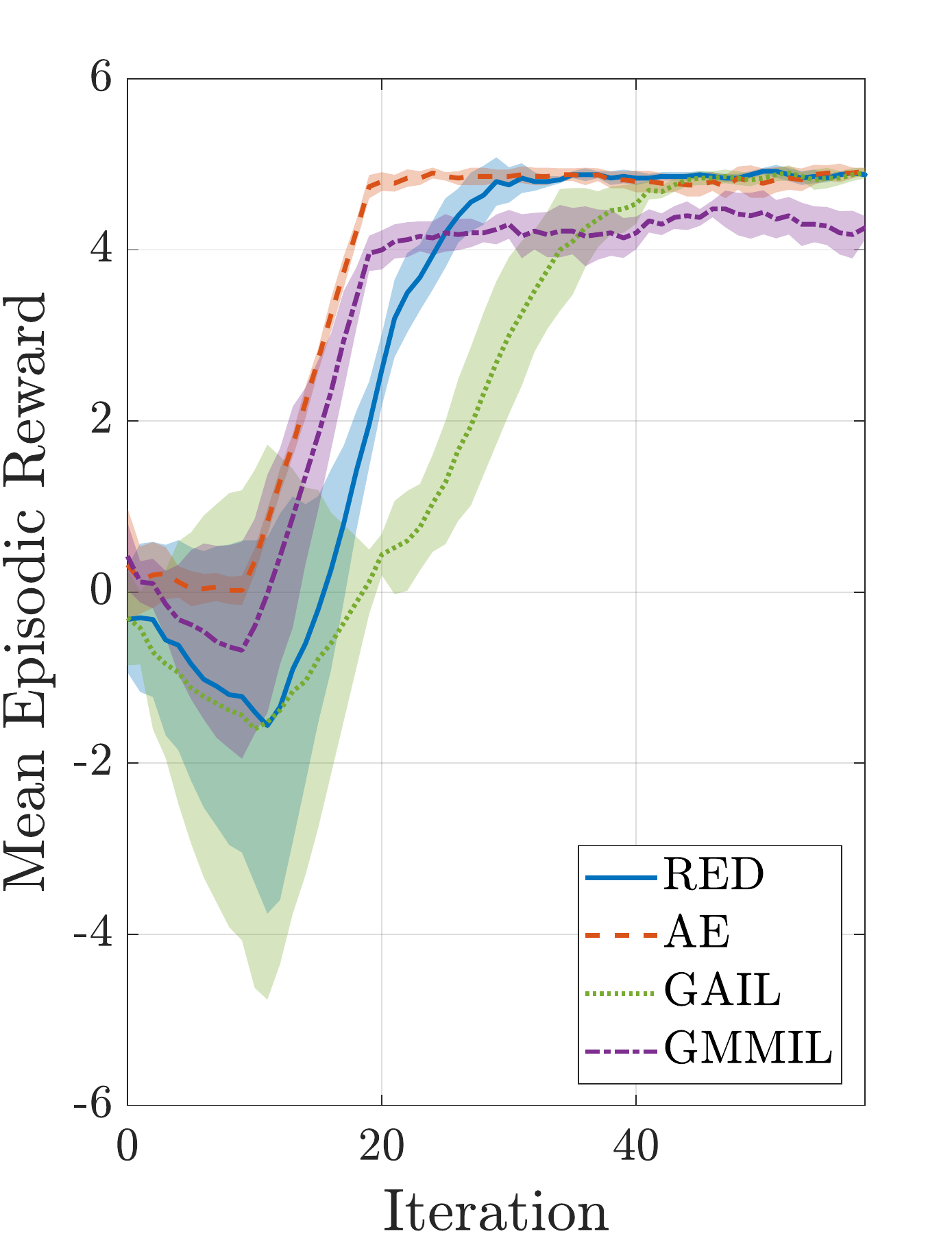}
        \caption{n = 50}
        \label{fig:toy_50}
    \end{subfigure}
    \begin{subfigure}[t]{.24\textwidth}
        \includegraphics[width=\textwidth]{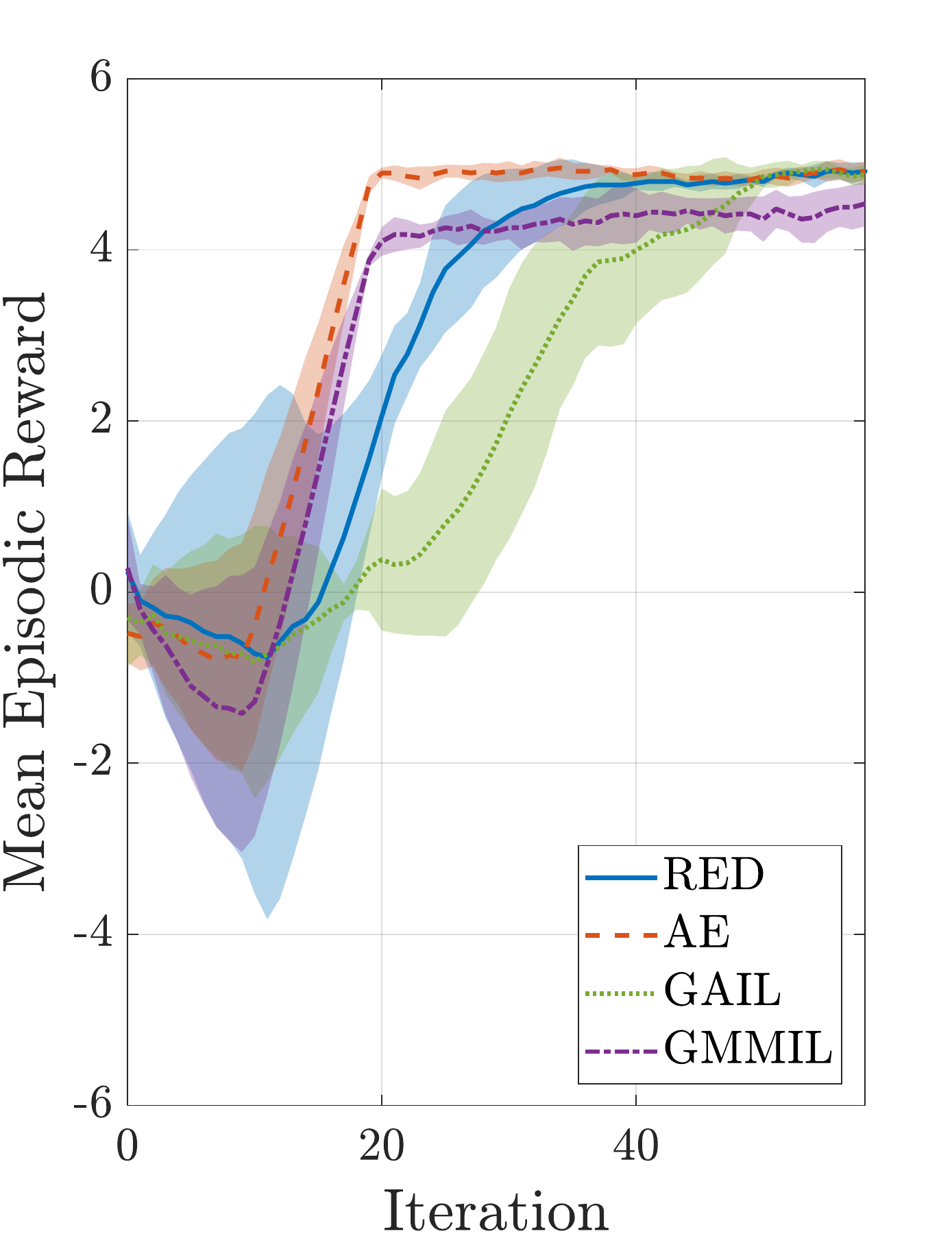}
        \caption{n = 100}
        \label{fig:toy_100}
    \end{subfigure}
    \caption{True mean episodic reward during training on the simple domain, with different expert dataset sizes.}
    \label{fig:toy_comp}
\end{figure*}

\section{Experiments}
We evaluate the proposed method on multiple domains. We present a toy problem to motivate the use of expert policy support estimation for imitation learning; and to highlight the behaviors of different imitation learning algorithms. We then evaluate the proposed reward function on five continuous control tasks from the Mujoco environment. Lastly, we test on an autonomous driving task with a single trajectory provided by a human driver. The code for reproducing the experiments is available online.\footnote{\url{https://github.com/RuohanW/red.git}}

\subsection{Simple Domain}\label{sec:toy}
We consider a stateless task where $s \sim \text{unif}(-1, 1)$, a discrete action space $a \in \{-1, 1\}$ and the reward function $r(s, a) = as$. It is clear that the expert policy to this problem is $\pi_E(s)=\text{sign}(s)$. Using Deep Q Learning \cite{mnih2015human} as the reinforcement learning algorithm, we compare the proposed method against AE, GAIL and GMMIL with an expert dataset of size $n=5, 10, 50, 100$ respectively.

\begin{figure}[t]
    \centering   
    \begin{subfigure}{.32\textwidth}
        \includegraphics[width=\textwidth]{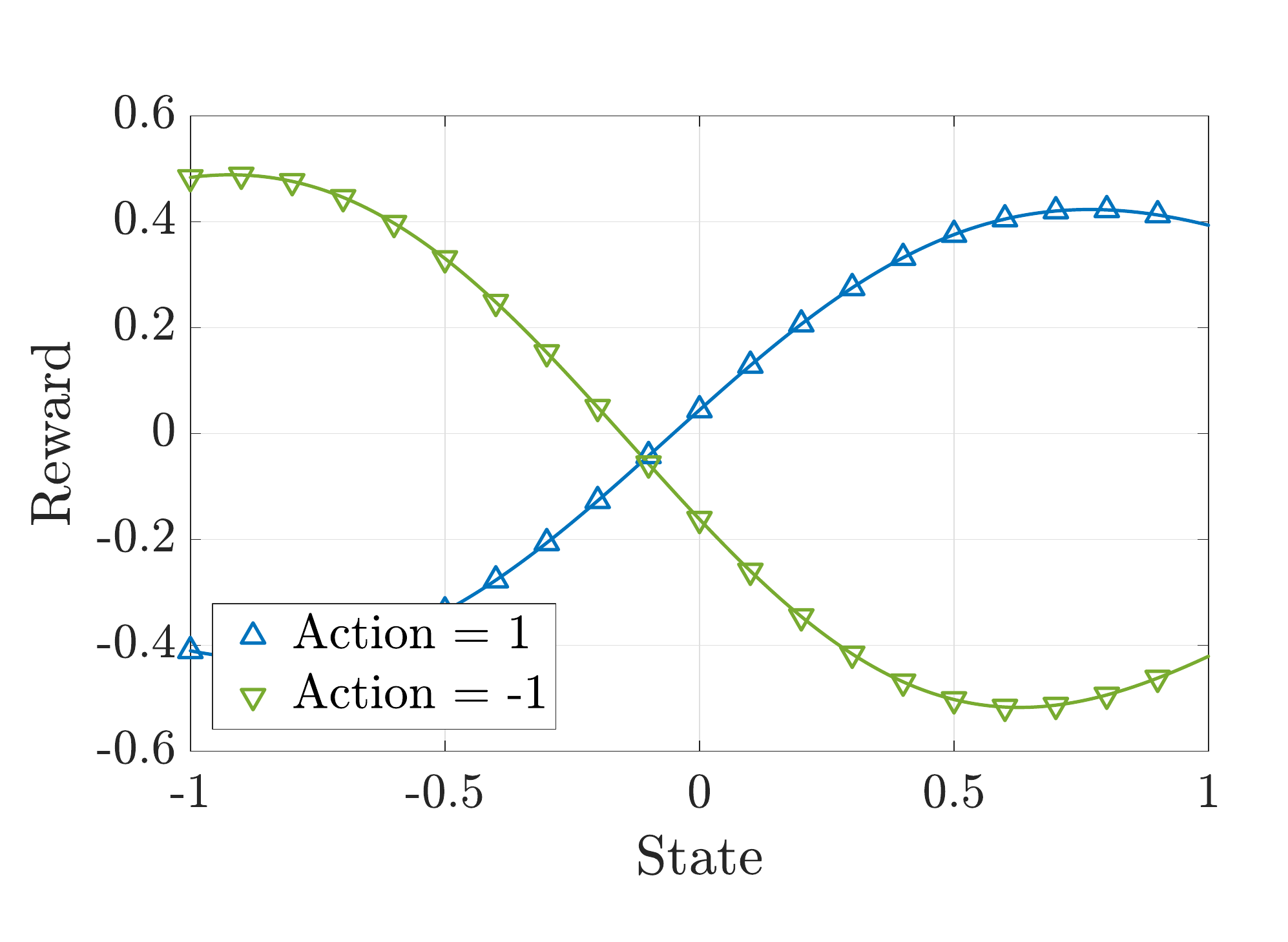}
        \caption{GMMIL}
        \label{fig:GMMIL_toy}
    \end{subfigure}
    \begin{subfigure}{.32\textwidth}
        \includegraphics[width=\textwidth]{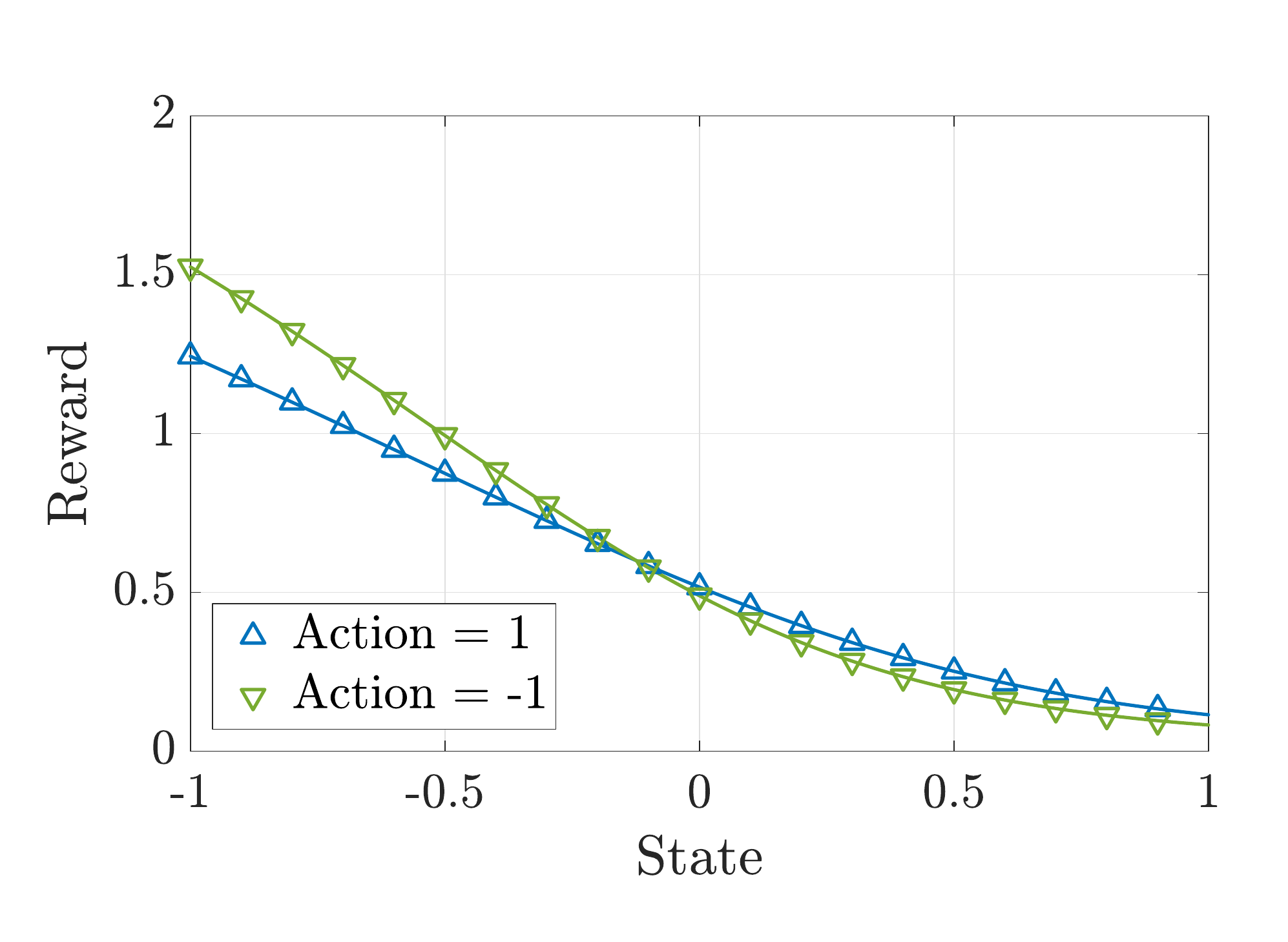}
        \caption{ GAIL}
        \label{fig:GAIL_toy}
    \end{subfigure}
    \begin{subfigure}{.32\textwidth}
        \includegraphics[width=\textwidth]{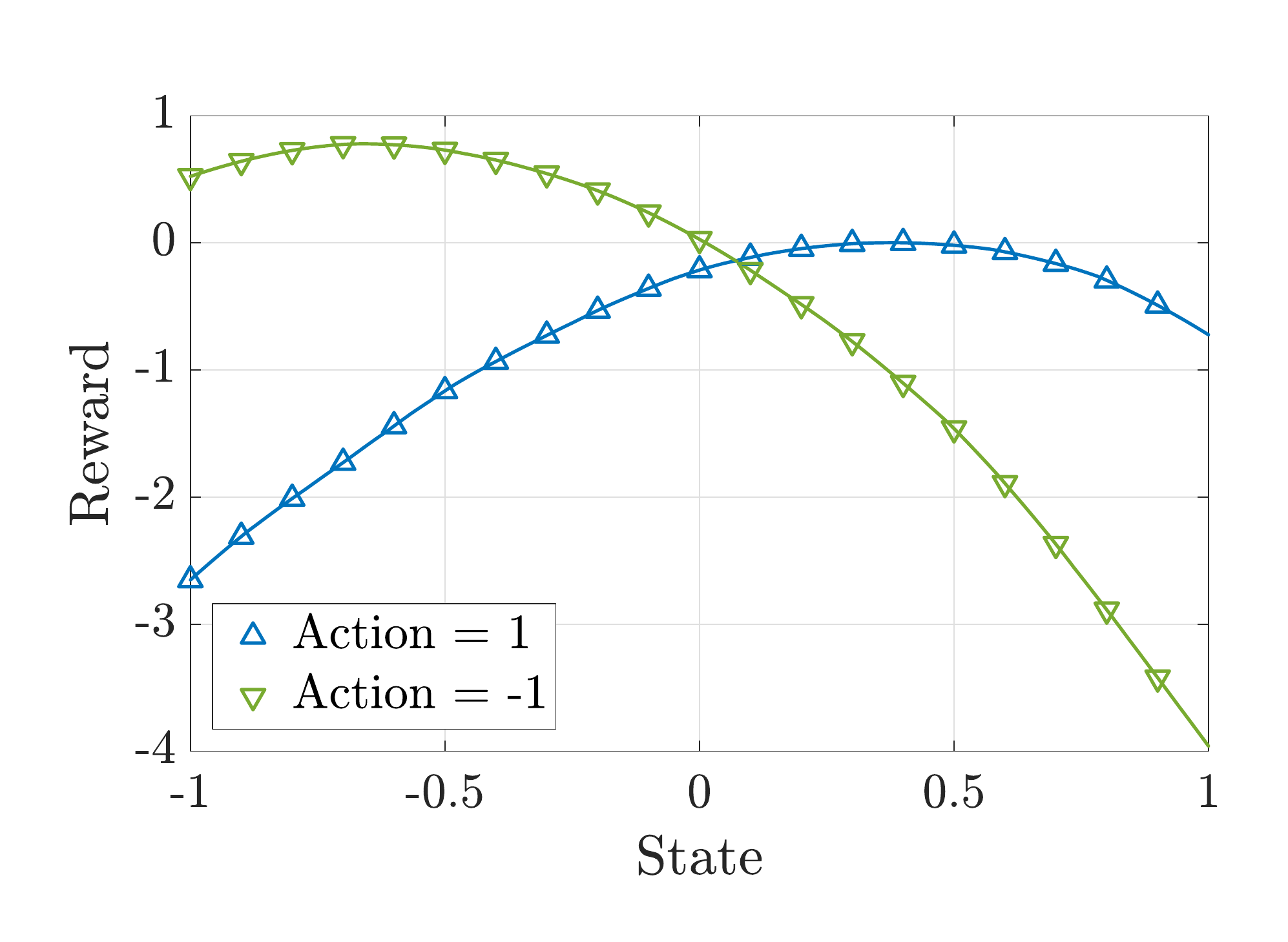}
        \caption{ AE}
        \label{fig:ENC_toy}
    \end{subfigure}
    
    \begin{subfigure}{.32\textwidth}
        \includegraphics[width=\textwidth]{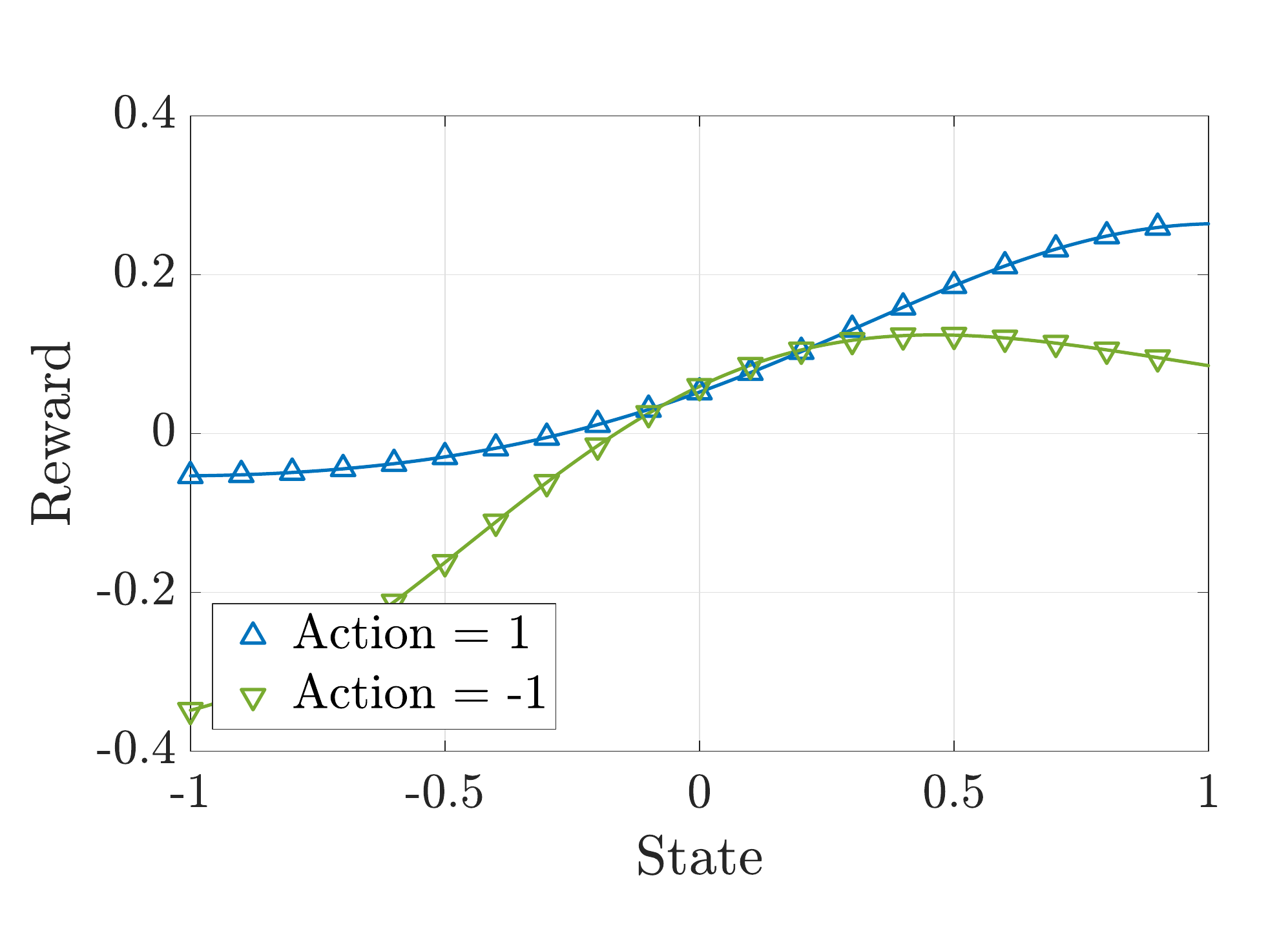}
        \caption{ GMMIL fails intermittently}
        \label{fig:GMMIL_toy_b}
    \end{subfigure}
    \begin{subfigure}{.32\textwidth}
        \includegraphics[width=\textwidth]{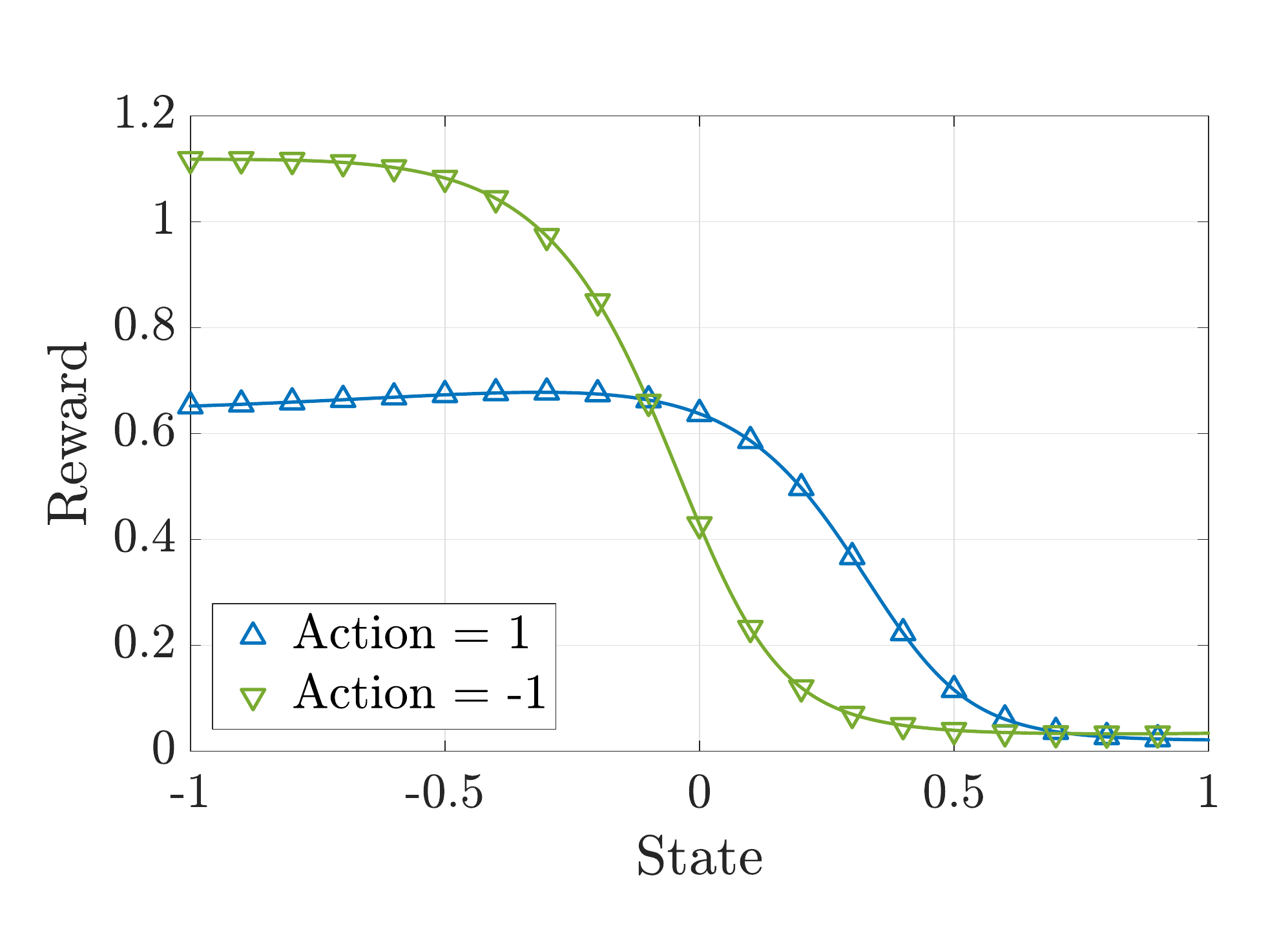}
        \caption{ \centering GAIL overfits with state near 1}
        \label{fig:GAIL_toy_b}
    \end{subfigure}
    \begin{subfigure}{.32\textwidth}
        \includegraphics[width=\textwidth]{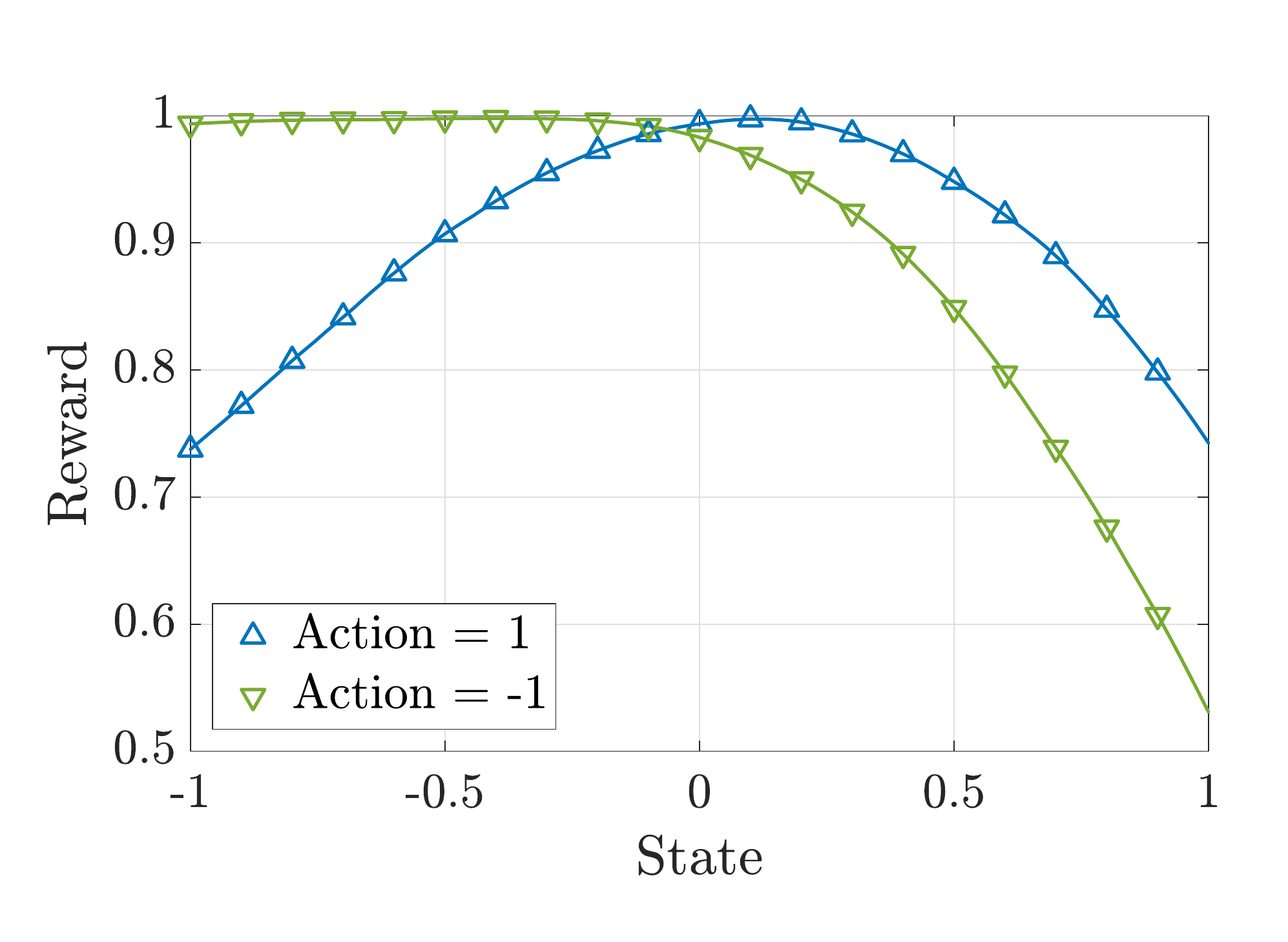}
        \caption{ RED }
        \label{fig:RED_toy}
    \end{subfigure}
    \caption{Estimated reward functions of the expert using different imitation learning algorithms on the simple domain. For better visualization of the reward, we use $r(s, a)=1 - \alpha_1 L(s, a)$ as the reward function for AE and RED.}\label{fig:toy_reward}
\end{figure}

In Figure \ref{fig:toy_comp}, we show the true mean episodic reward of different algorithms during training. Except for GMMIL, all other algorithms converge to the expert policy for all sizes of the dataset. In addition, RED and AE converge to the expert policy faster as the extracted reward functions recover the correct reward function nearly perfectly (Figures \ref{fig:RED_toy}, \ref{fig:ENC_toy}). In contrast, GAIL requires adversarial training to recover the correct reward function, a noisy process during which all actions from the learned policy, both optimal or not, are considered ``wrong'' by the discriminator. In fact, when the discriminator is overparametrized, it would overfit to the training data gradually and generate arbitrary reward for some region of the state-action space (Figure \ref{fig:GAIL_toy_b}). The results are consistent with observations from \cite{brock2018large} that the discriminator is overfitting, and that early stopping may be necessary to prevent training collapse. For GMMIL, we observe that the method intermittently generates wrong reward functions (Figure \ref{fig:GMMIL_toy_b}), causing the performance to oscillate and converges to lower mean reward, especially when the expert data is sparse. This is likely due to the noise in estimating distribution moments from limited samples.



\begin{table*}[t]
\vskip 0.15in
\caption{Episodic reward (as provided by the environment) on Mujoco tasks by different methods evaluated over 50 runs. HalfCheetah and Ant uses BC initialization in RED and AE. We are unsuccessful with GMMIL on Ant.}
\begin{center}
\begin{scriptsize}
\begin{sc}
\begin{tabular}{cccccc}
\toprule
& Hopper & HalfCheetah & Reacher & Walker2d & Ant\\
\midrule
GAIL & 3614.22 $\pm$ 7.17 & 4515.70 $\pm$ 549.49 & -32.37 $\pm$ 39.81 & 4877.98 $\pm$ 2848.37 & 3186.80 $\pm$ 903.57\\
GMMIL & 3309.30 $\pm$ 26.28 & 3464.18 $\pm$ 476.50 & -11.89 $\pm$ 5.27 & 2967.10 $\pm$ 702.03 & -\\
AE & 3478.31 $\pm$ 3.09 & 3380.74 $\pm$ 101.94 & -10.91 $\pm$ 5.62 & 4097.61 $\pm$ 118.06 & 3778.61 $\pm$ 422.63\\
RED & 3625.96 $\pm$ 4.32 & 3072.04 $\pm$ 84.71 & -10.43 $\pm$ 5.20 & 4481.37 $\pm$ 20.94 & 3552.77 $\pm$ 348.67\\
\bottomrule
\end{tabular}

\label{tab:mujoco}
\end{sc}
\end{scriptsize}
\end{center}
\vskip -0.1in
\end{table*}

\subsection{Mujoco Tasks}
We further evaluate RED on five continuous control tasks from the Mujoco environment: Hopper, Reacher and HalfCheetah, Walker2d and Ant. Similarly, we compare RED against AE, GAIL and GMMIL, using Trust Region Policy Optimization (TRPO) \cite{schulman2015trust} in Table \ref{tab:mujoco}. Similar to the experiments in \cite{ho2016generative}, we consider learning with 4 trajectories of expert demonstration generated by an expert policy trained with RL. All RL algorithms terminate within 5M environment steps. We note that we were unsuccessful in the Ant task with GMMIL after an extensive search for the appropriate hyperparameters.


\begin{figure}[t]
\centering
\includegraphics[width=.50\textwidth, trim={0 0 0 .0cm}, clip]{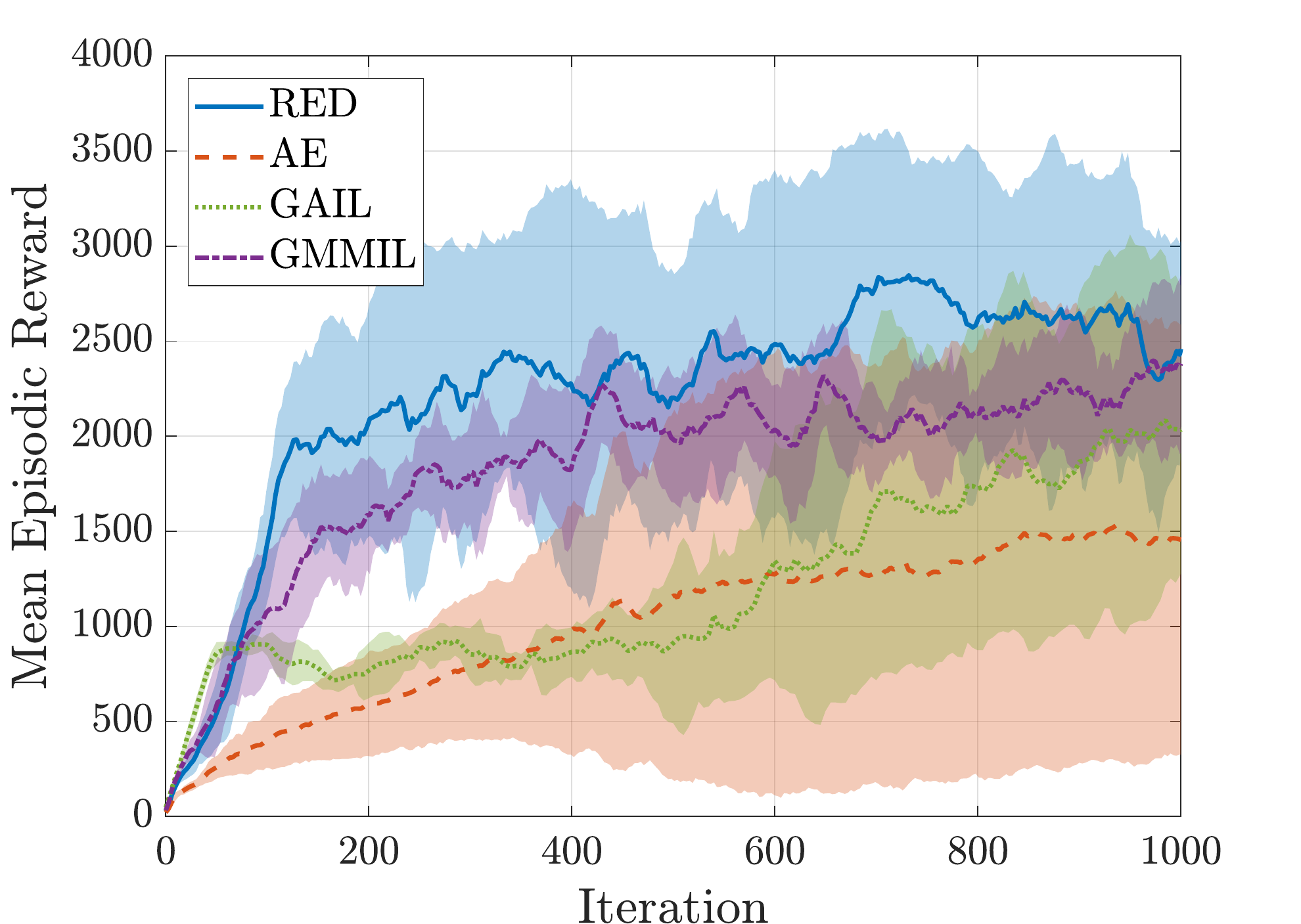}
\caption{True mean episodic reward of GMMIL, GAIL, AE and RED on Hopper during training. Agents trained with RED improve faster compared to other methods.}
\label{fig:trpo}
\end{figure}

The results suggest that support estimation of expert policies is a viable strategy for imitation learning, as the AE and RED are both able to achieve good results with the fixed reward functions constructed from support estimation of the expert policy. While RED and AE underperform on the HalfCheetah, both methods are comparable or better than GAIL and GMMIL in all other tasks. In particular, RED and AE achieve much smaller variance in performance, likely due to the fixed reward function. Consistent with the observations from the simple domain in Sec.~\ref{sec:toy}, RED appears to achieve faster performance improvements during early training (Figure \ref{fig:trpo}).  We note that though the best performance AE can achieve is similar to RED, AE is much more sensitive to the random initialization of parameters, seen from the large standard deviation in Figure \ref{fig:trpo}.

A limitation of our method is that HalfCheetah and Ant require a policy initialized with BC to achieve good results while GAIL and GMMIL could start with a random policy in the two tasks. We hypothesize that the evolving reward functions of GAIL and GMMIL may provide better exploration incentives to the RL agent. As GAIL is orthogonal to our method and may be used together, we leave it to future work on how to combine the benefits of both RED and GAIL to achieve more robust algorithms.

\begin{figure}[t]
    \centering
    \begin{subfigure}{.4\textwidth}
    \caption{Episodic reward (distance travelled without collision) on the driving task with different methods. The expert performance of 7485 corresponds with track completion without collision. Both AE and RED uses the terminal reward heuristic.}
    \label{tab:car}
    \vskip 0.15in
    \begin{center}
    \begin{footnotesize}
    \begin{sc}
    \begin{tabular}{cccc}
    \toprule
    & Average & Std & Best\\
    \midrule
    GAIL & 795 & 395 & 1576\\
    GMMIL & 2024 & 981 & 3624\\
    BC & 1033 & 474 & 1956\\
    AE & 4378 & 1726 & 7485\\
    RED & 4825 & 1552 & 7485\\
    Expert & 7485 & 0 & 7485\\ 
    \bottomrule
    \end{tabular}
    \end{sc}
    \end{footnotesize}
    \end{center}
    \vskip -0.1in
    \end{subfigure}
    \qquad 
    \begin{subfigure}{.4\textwidth}
    \centering
    \includegraphics[width=0.9\textwidth, trim={0 0 0 5cm}, clip]{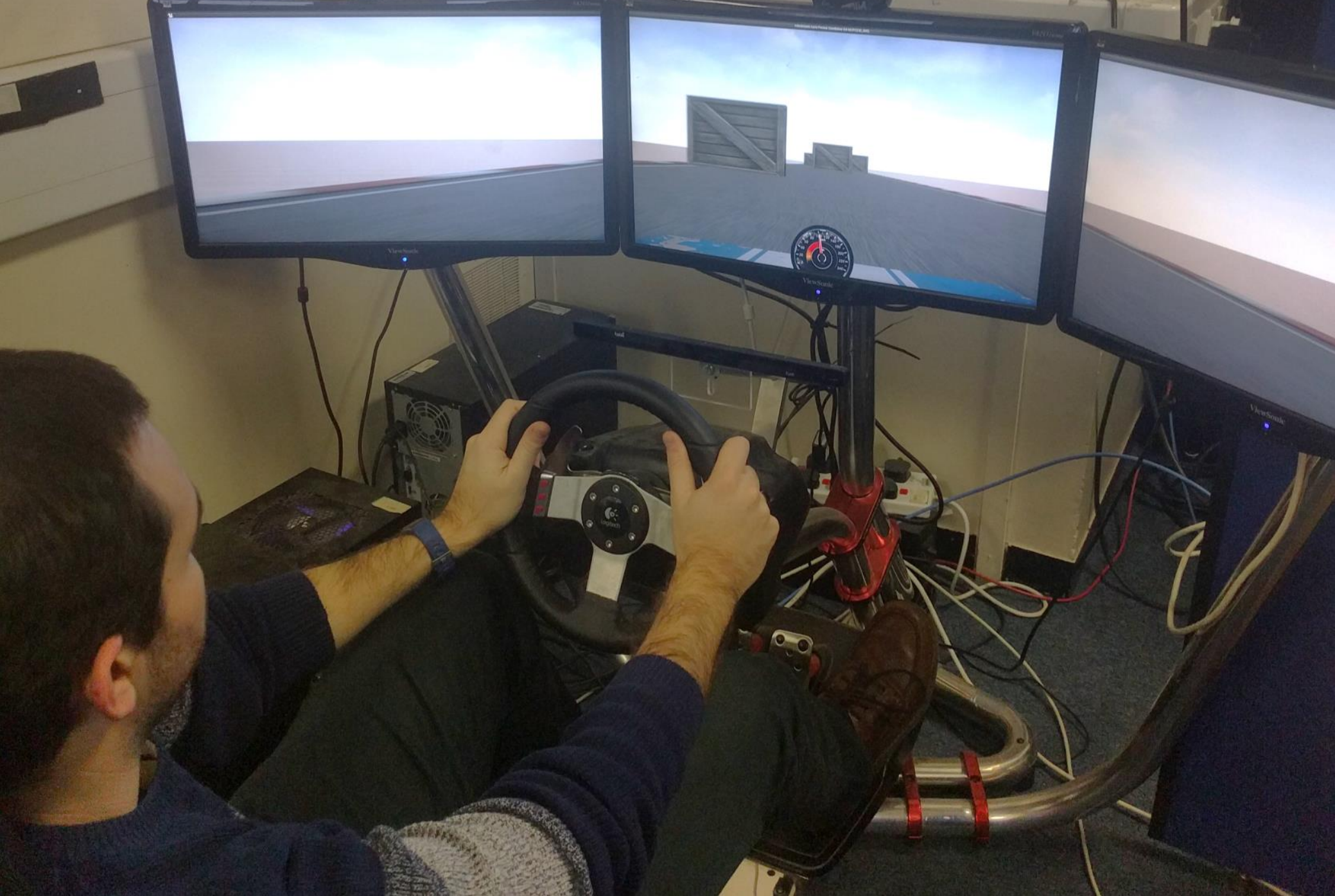}
    \caption{The expert driver providing demonstration on the driving scenario. The scenario consists of driving through a straight track while avoiding obstacles.}
    \label{fig:setup}
    \end{subfigure}
\end{figure}

\subsection{Autonomous Driving Task}
Lastly, we evaluate RED, AE, GAIL, GMMIL and BC on an autonomous driving task, using a single demonstration provided by a human driver. The environment consists of a straight track with obstacles placed randomly at one of the three lanes of the track (Figure \ref{fig:setup}). A human demonstrator was instructed to drive through the track and avoid any obstacles, while keeping the speed around 100 km/h. We sampled the expert driving actions at 20 Hz. For the environment, we use a vector of size 24 to represent state (20 dimensions for the LIDAR reading, 3 dimensions for the relative position, and orientation of the track with respect to the car, and 1 dimension for the vehicle speed) to output the steering and the accelerator command for the vehicle. We include the terminal reward heuristic defined in Eq. \ref{eq:term_rew}. For evaluation, we measure the distance a learned policy is able to control the vehicle through the track without any collision with the obstacles. The task is challenging as the learner must generalize from the limited amount of training data, without explicit knowledge about the track width or the presence of obstacles. The driving task also allows us to qualitatively observe the behaviors of learned policies.

In this task, we initialize all policies with BC and use stochastic value gradient method with experience replay \cite{heess2015learning} as the reinforcement learning algorithm. The algorithm is referred to as SVG(1)-ER in the original paper.


\begin{figure*}[t]
    \centering
    \begin{subfigure}[t]{.24\textwidth}
        \includegraphics[width=\textwidth]{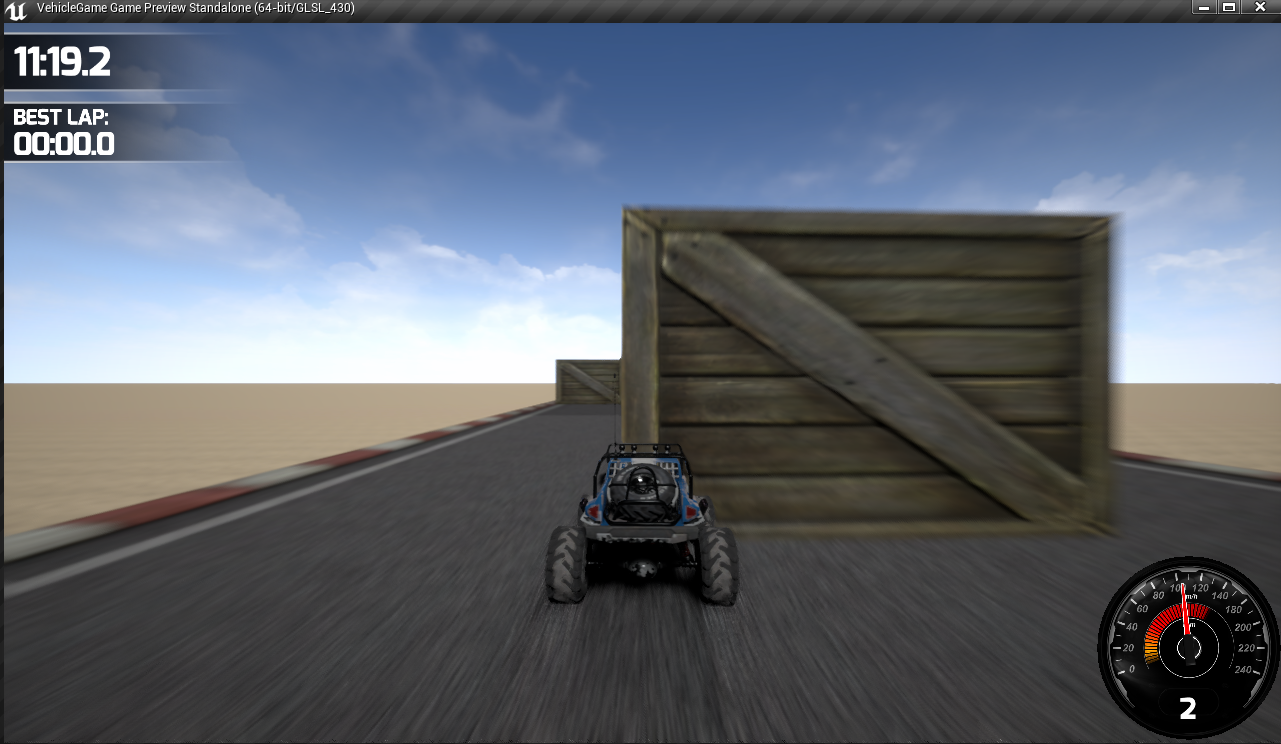}
        \caption{Near Collision}
    \end{subfigure}
    \begin{subfigure}[t]{.24\textwidth}
        \includegraphics[width=\textwidth]{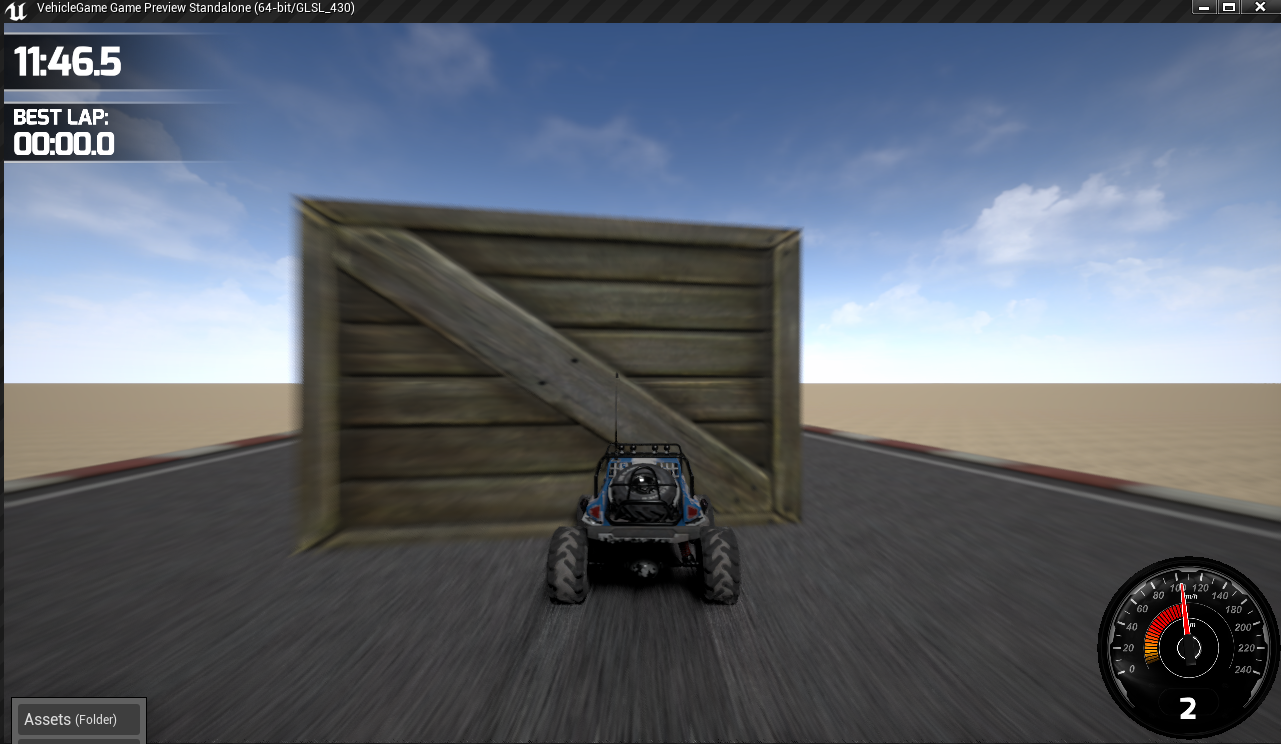}
        \caption{Collision}
    \end{subfigure}
    \begin{subfigure}[t]{.24\textwidth}
        \includegraphics[width=\textwidth]{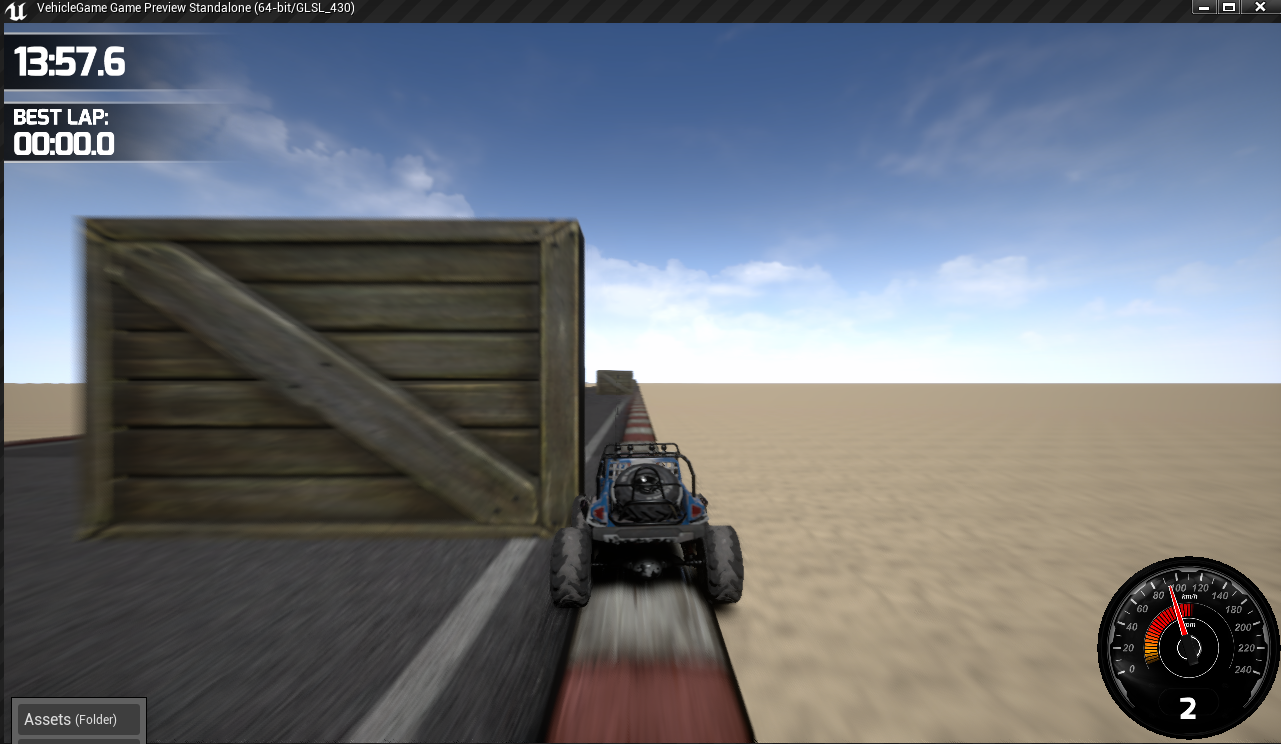}
        \caption{Off-road Collision}
    \end{subfigure}
    \begin{subfigure}[t]{.24\textwidth}
        \includegraphics[width=\textwidth]{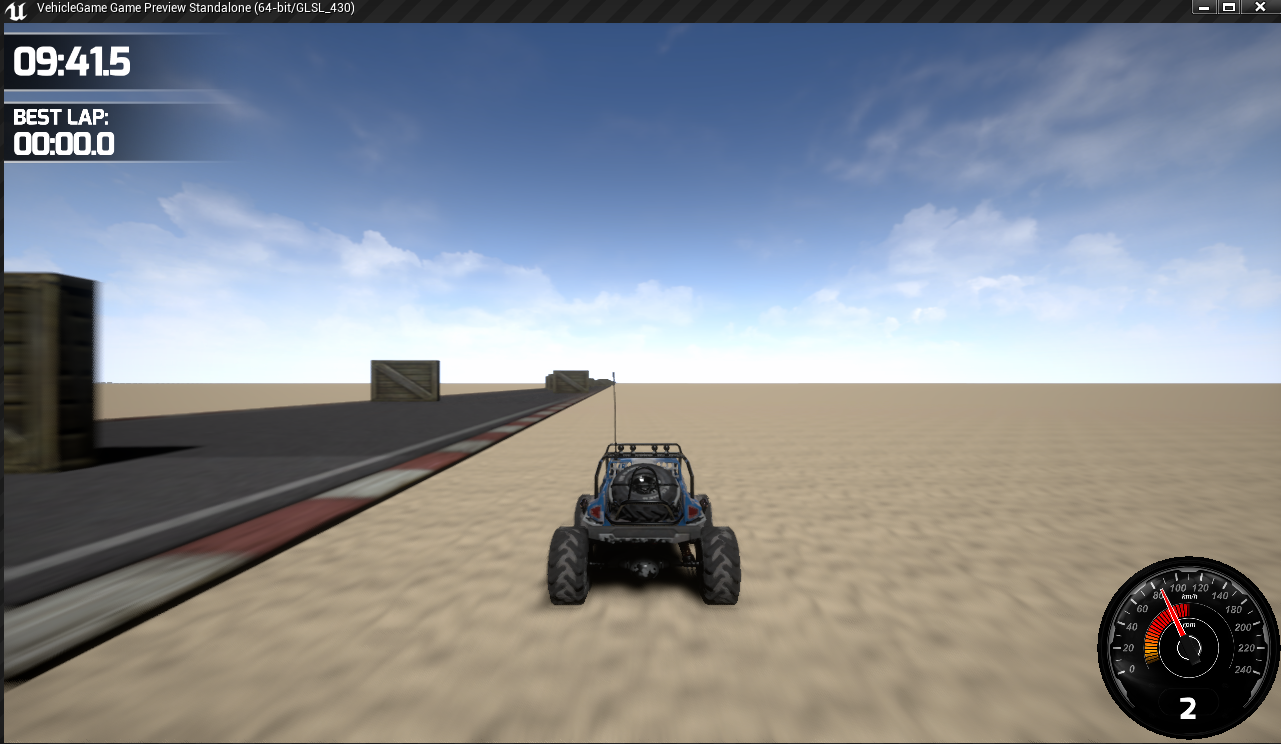}
        \caption{Off-road}
    \end{subfigure}
    \caption{Representative scenarios where the reward functions of AE and RED assign near zero rewards. The scenarios correspond well with various dangerous states as they are dissimilar to those from expert demonstrations.}\label{fig:car_danger}
\end{figure*}

Table \ref{tab:car} shows the average and best performance of each method evaluated over 5 runs through the same track. We note that the expert performance of 7485 corresponds with track completion without collision. The results suggest that RED achieves the best performance. For both RED and AE, the reward functions correctly identify dangerous situations, such as collision or near-collision, by assigning those states with zero rewards. Figure \ref{fig:car_danger} shows a few representative states where the reward function outputs zero reward. In contrast, we observe that GAIL tends to overfit to the expert trajectory. During training, GAIL often "forgets" how to avoid the same obstacle after the discriminator update the reward function. Such behaviors prevent the learned policy from avoiding obstacles consistently. For GMMIL, we observe behaviors similar to that found in Sec.~\ref{sec:toy}: the reward function intermittently produces problematic incentives, such as assigning positive rewards for states immediately preceding collisions.

\section{Conclusion}
We propose a new general framework of imitation learning via expert policy support estimation. We connect techniques such as Random Network Distillation and AutoEncoders to approximate support estimation; and introduce a method for efficiently learning a reward function suitable for imitation learning from the expert demonstrations. We have shown empirically in multiple tasks that support estimation of expert policies is a viable strategy for imitation learning, and achieves comparable or better performance compared to the state-of-art. For future works, combining different approaches of recovering the expert's reward function appears to be a promising direction.
\bibliography{main}
\bibliographystyle{icml2019}

\end{document}